\documentclass[10pt,twocolumn,letterpaper]{article}

\usepackage[pagenumbers]{cvpr} % To force page numbers, e.g. for an arXiv version

\usepackage{graphicx}
\usepackage{amsmath}
\usepackage{amssymb}
\usepackage{booktabs}

\usepackage[pagebackref,breaklinks,colorlinks]{hyperref}

\usepackage[capitalize]{cleveref}
\usepackage{multirow}
\usepackage{colortbl}
\usepackage{graphicx}
\usepackage{grffile}
\usepackage[ruled,linesnumbered]{algorithm2e}
\usepackage{amsmath}
\usepackage{booktabs} % For formal tables
\usepackage{color}
\usepackage{subfloat}
\usepackage{balance}
\usepackage{multicol}
\usepackage{bbding}
\usepackage{makecell}
\usepackage{amstext}
\usepackage{setspace}
\usepackage[figuresright]{rotating}
\usepackage{lscape}
\usepackage{adjustbox}
\usepackage{capt-of}
\usepackage{enumerate}
\usepackage{makecell}
\usepackage{amssymb}
\crefname{section}{Sec.}{Secs.}
\Crefname{section}{Section}{Sections}
\Crefname{table}{Table}{Tables}
\crefname{table}{Tab.}{Tabs.}

\def\cvprPaperID{6727} % *** Enter the CVPR Paper ID here
\def\confName{CVPR}
\def\confYear{2022}

\begin{document}

\title{Non-generative Generalized Zero-shot Learning via Task-correlated Disentanglement and Controllable Samples Synthesis}

\author{Yaogong Feng, 
Xiaowen Huang\footnotemark[1],
Pengbo Yang, 
Jian Yu,
Jitao Sang\\
School of Computer and Information Technology \\
\& Beijing Key Lab of Traffic Data Analysis and Mining, Beijing Jiaotong University, China\\
% $^1$School of Computer and Information Technology, Beijing Jiaotong University, China\\
% $^2$Institute of artificial intelligence, Beijing Jiaotong University, China\\
% Beijing, China\\
{\tt\small \{fengyg18, xwhuang, pengboyang, jianyu, jtsang\}@bjtu.edu.cn}
% \and
}
\maketitle

\renewcommand{\thefootnote}{\fnsymbol{footnote}} 
\footnotetext[1]{Corresponding authors.}

%%%%%%%%% ABSTRACT
\begin{abstract}
Synthesizing pseudo samples is currently the most effective way to solve the Generalized Zero-Shot Learning (GZSL) problem. Most models achieve competitive performance but still suffer from two problems: (1) \textbf{Feature confounding}, the overall representations confound task-correlated and task-independent features, and existing models disentangle them in a generative way, but they are unreasonable to synthesize reliable pseudo samples with limited samples; (2) \textbf{Distribution uncertainty}, that massive data is needed when existing models synthesize samples from the uncertain distribution, which causes poor performance in limited samples of seen classes. 
In this paper, we propose a non-generative model to address these problems correspondingly in two modules: (1) \textbf{Task-correlated feature disentanglement}, to exclude the task-correlated features from task-independent ones by adversarial learning of domain adaption towards reasonable synthesis; (2) \textbf{Controllable pseudo sample synthesis}, to synthesize edge-pseudo and center-pseudo samples with certain characteristics towards more diversity generated and intuitive transfer. In addation, to describe the new scene that is the limit seen class samples in the training process, we further formulate a new ZSL task named the 'Few-shot Seen class and Zero-shot Unseen class learning' (FSZU). Extensive experiments on four benchmarks verify that the proposed method is competitive in the GZSL and the FSZU tasks. 

% Synthesizing pseudo samples is currently the most effective way to solve the Generalized Zero Shot Learning (GZSL) problem. Most models achieve competitive performance but still suffer from two problems: (1) feature confounding, that task-correlated and task-independent features are confounded and unreasonably contribute to the knowledge transfer from seen to unseen classes; (2) distribution uncertainty, the pseudo samples are typically synthesized from the uncertain distribution of unseen classes in existing models. In this paper, we propose a non-generative GZSL method to address these two problems correspondingly in two modules: (1) task-correlated feature disentanglement, to separate the task-correlated features from task-independent ones by adversarial learning of domain adaption towards reasonable transfer; and (2) controllable pseudo sample synthesis, to synthesize edge-pseudo and center-pseudo samples respectively towards more interpretable transfer. In addition to the existing ZSL tasks, we further propose a new ZSL task named the 'Few-shot Seen class and Zero-shot Unseen class learning'. Quantitative as well as qualitative experiments on four benchmarks verify that the proposed method is competitive with recent GSZL methods in both efficiency and interpretability. 
\end{abstract}

%%%%%%%%% BODY TEXT
\section{Introduction}
\label{sec:intro}

The explosion of data and the rapid development of deep learning need massive precise but expensive labels. In the real world, these labels are usually sparse/missing, Zero-Shot Learning (ZSL) techniques offer a good solution to address such problem, which trains on seen classes and tests on unseen classes (the seen classes and unseen classes are independent). In this paper, we focus on the Generalized ZSL (GZSL) task. It is a realistic setting of ZSL in making predictions on recognizing samples from both classes simultaneously rather than classifying only data samples of the unseen classes.
\begin{figure}
  \centering
  \includegraphics[width=80mm]{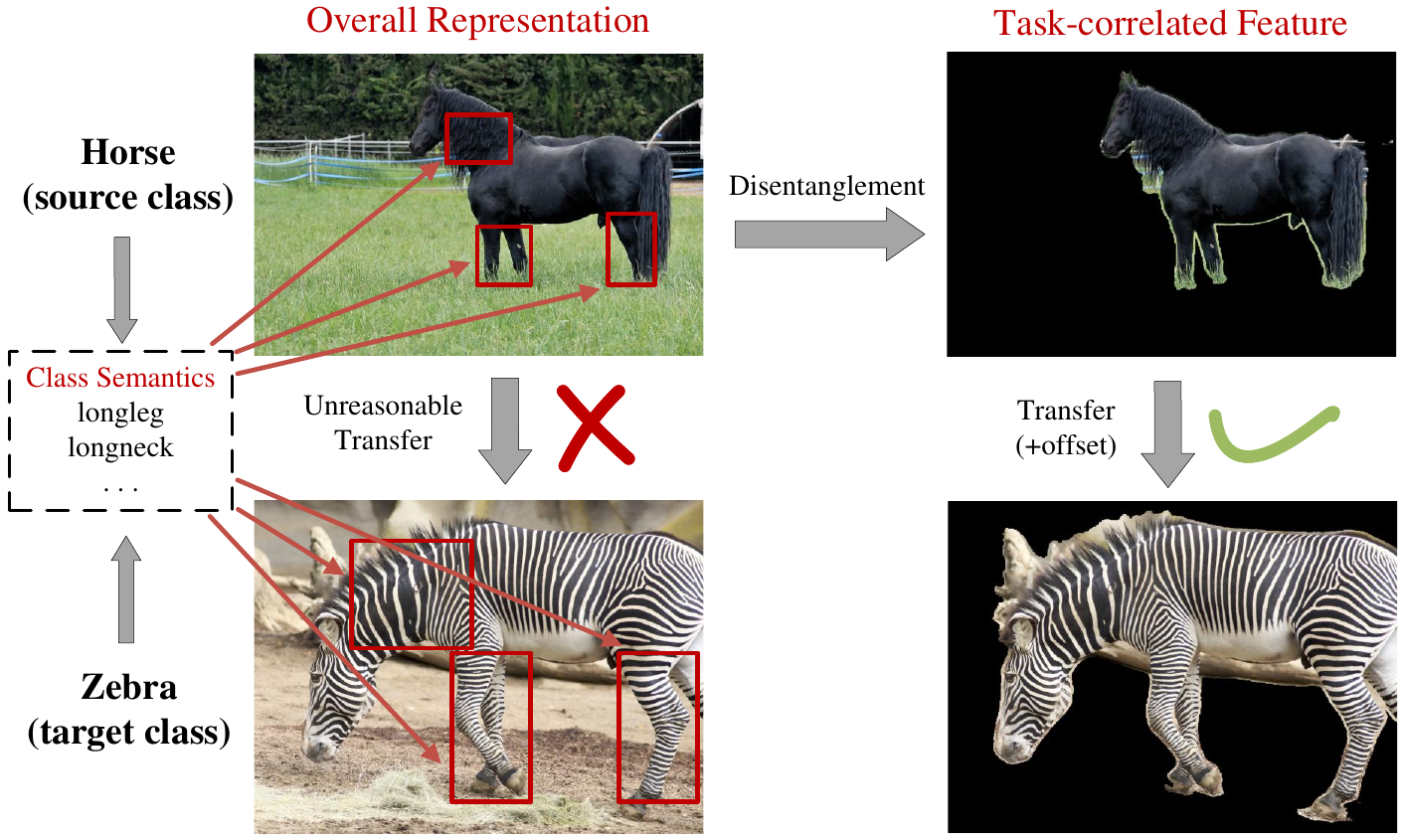}\\
  \caption{An illustration of core idea of our method. Different classes share class semantics. The images in the right half are the task-correlated features of the left half. They are excluded from task-independent features and are more consistent with class semantics, which is reasonable for the knowledge transfer in the GZSL task.}
%   The transfer process between horse class and zebra class in the right half is more reasonable than the one in the left half for the GZSL task.
  \label{fig:1}
  \vspace{-0.5cm}
\end{figure}

At present, the method of synthesizing pseudo samples for unseen classes has proven to be one of the most effective ways of knowledge transfer to solve the GZSL problem. But there are still two challenging problems:
(1) \textbf{Feature Confounding.} Most GZSL models are based on overall representations of samples extracted in pre-trained CNN (e.g., ResNet101 \cite{he2016deep}) while the semantic features are class-level attributes or class-level sentence embeddings \cite{lampert2013attribute,wah2011caltech}. The former contains more rich information and inconsistent with human cognition. So it is unreasonable to construct a mapping from visual features to semantic features directly and synthesize reliable pseudo samples based on the confounding visual features. 
Although some models \cite{li2021generalized,chen2021semantics} contribute to extracting more human-consistent-cognition features, they are related to generative models and hard to guarantee the diversity of sample synthesis with limited real samples.

(2) \textbf{Distribution Uncertainty.} The existed methods, especially the generative models, usually require a large amount of data to fit the distribution of real data, and can only synthesize pseudo samples with the uncertain distribution. So these models have poor performances when samples of seen class are few shot.

Figure \ref{fig:1} illustrates the motivation of our method. 
% Different classes share items of class semantic, such as legs, necks, and so on. 
We first exclude the task-independent feature from horse image to horse object. Then we synthesize zebra's pseudo samples of target class based on the horse's task-correlated feature of source class in a non-generative way.

To be specific, we propose a non-generative approach named Task-correlated Disentanglement and Controllable Samples Synthesis (TDCSS) method that handles above issues. The TDCSS mainly consists of two components.
(1) \textbf{Task-correlated Feature Disentanglement Module.} Our model is based on class semantic features to complete the image classification task. According to whether the visual features are corresponding to class semantics, we disentangle the confounding features into task-correlated features and task-independent features. The task-correlated features are more consistent with class semantics.
We introduce the adversarial training of domain adaptation to achieve feature disentanglement.
(2) \textbf{Controllable Pseudo Samples Synthesis Module.} Based on task-correlated features, we add different offsets to synthesize two types of pseudo samples which are edge-pseudo samples and center-pseudo samples in a non-generative way. For the edge-pseudo samples, we treat them as the adversarial examples in the feature-level and the edge offsets can be seen as perturbations in adversarial examples. For the center-pseudo samples, we make them distributed closer to the center of one class. Both of them guarantee the generative diversity of samples based on the limited seen samples. In addition, synthesizing pseudo samples with certain characteristics contribute to exploring the role of different types of pseudo samples in the knowledge transfer of GZSL.

To describe the scene that only has the limited samples on seen class in formulation, we further propose a new ZSL task named 'Few-shot Seen class and Zero-shot Unseen class learning (FSZU)'. The FSZU is also more reasonable and more practical. In the GZSL task, all classes have strong semantic relationships. So we believe that the ZSL and the Few-Shot Learning (FSL) are coexistent. For example, in the deep-space exploration and deep-sea exploration, the machine (detector) always encounter new situations, and the number of seen class samples that humans have obtained is also extremely limited. In this paper, we perform the TDCSS and the similar methods on the new task.

In summary, our main contributions of this work are summarized as follows:

(1)	We propose a novel non-generative model that disentangles visual features into task-correlated and task-independent by adversarial training of domain adaptation. And we use the task-correlated features to synthesize two types of pseudo samples as center-pseudo samples and edge-pseudo samples to guarantee the diversity of sample synthesis and the intuitive transfer. 

(2)	We propose a new zero-shot task named 'Few-shot Seen class and Zero-shot Unseen class learning (FSZU)', which is more reasonable and more practical compared with the GZSL task in real life.

(3) Extensive experiments in the GZSL task and the FSZU task on four widely used datasets verify the results of the TDCSS are competitive with similar methods.

\section{Related Works}
\subsection{Generalized Zero-Shot Learning}
Domain shift is a basic problem in GZSL. It is described as 'due to having disjoint and potentially unrelated classes, the projection functions learned from the auxiliary dataset/domain are biased when applied directly to the target dataset/domain' \cite{fu2017zero}. 

\textbf{GZSL with pseudo sample synthesis.} Many researchers have proposed generative models to synthesize pseudo samples for unseen classes to alleviate this problem. 
The GAN-based models contribute to increasing the diversity of pseudo samples \cite{li2019leveraging} and preserving semantic consistency \cite{ni2019dual, liu2020zero}. 
The VAE-based models \cite{schonfeld2019generalized, chen2020boundary, keshari2020generalized} contribute to preserving semantic consistency of different representations of distribution in hidden layer. 
Some researchers integrate the VAE and GAN into a unified conditional feature generating model \cite{xian2019f,narayan2020latent} to integrate the advantages.
There are also some non-generative models \cite{long2017zero, guo2017synthesizing, lu2018attribute, huynh2020compositional, guan2020zero, chou2020adaptive} synthesizing pseudo samples. The models \cite{lu2018attribute,huynh2020compositional} extract attribute-based features and then combine them to synthesize unseen pseudo samples. The BPL \cite{guan2020zero} synthesizes pseudo samples based on bidirectional projection learning and linear interpolation. The AGZSL \cite{chou2020adaptive} uses the Image Adaptive Semantics to expand the semantic features by using visual features, then it trains the seen class classifier based on the expanded semantic features and trains the unseen class classifier by synthesizing virtual class by sampling and interpolating over seen counterparts. The models \cite{guan2020zero,chou2020adaptive}, synthesizing pseudo samples by perturbating or interpolating, are simpler and more likely to transfer class variations from seen class to unseen classes. So we use the non-generative models to synthesize pseudo samples.

\textbf{GZSL with representation disentanglement.} Most of GZSL models are based on the overall representations. Researchers use the representation disentanglement to obtain more consistent visual features with semantics. 
Some researchers disentangle visual features into 'object + attribute' features \cite{misra2017red,atzmon2020causal} by exploring their respective distributions. For example, the 'red wine' can be disentangled into 'red (attribute)' + 'wine (object)'. But these methods require more strictly labeled datasets.
Some researchers try to align attribute-based features with their attribute semantic vectors in fine-grained ZSL task \cite{huynh2020fine,huynh2020compositional}. But their methods are based on the feature map at the last convolutional layer.
Some researchers disentangle visual features according to their understanding of the GZSL. The DLFZRL \cite{tong2019hierarchical} disentangles the feature into the semantic latent feature, the non-semantic latent feature, and the non-discriminative latent feature, in which the first two factors are learned by adversarial learning and the last is learned by a hierarchical structure. In addition, SP-AEN \cite{chen2018zero} disentangles the semantic space into two subspaces for classification and reconstruction respectively. Some researchers also use generative models with random permuting to achieve representation disentanglement \cite{li2021generalized,chen2021semantics}.
In this paper, we disentangle visual features into task-correlated features and task-independent features by using domain adversarial, which is more robust in different tasks.

\subsection{Adversarial Example and Adversarial Self-Supervised Learning}
Recently, extensive experiments \cite{aggarwal2020benefits,shafahi2019adversarially,salman2020adversarially,utrera2020adversarially} have shown that model would have better generalization which has better adversarial robustness. And it achieves higher performance than the naturally trained models in ZSL \cite{aggarwal2020benefits}. These works usually use the gradient-based adversarial example generation algorithm, such as FSGM \cite{goodfellow2014explaining} and PGD \cite{kannan2018adversarial}. Some self-supervised learning methods \cite{kim2020adversarial,ho2020contrastive} are also upgraded to adversarial self-supervised learning based on the adversarial examples in the way of contrastive learning, which extracts image features that are more consistent with human cognition.

In this paper, we draw on that core idea and further propose the edge-pseudo samples that can be seen as feature-level adversarial examples based on the targeted attack. We also introduce a training mechanism of the adversarial self-supervised to TDCSS to make our model extract more consistent task-correlated features with class semantics. 

%-------------------------------------------------------------------------
\section{The Proposed Method}
\subsection{Problem Formulation}
In the GZSL task, let $\{ {{\cal X}^s},{{\cal Y}^s}\}$  be the dataset with  $S$ seen classes, which contains ${N^s}$ training samples ${{\cal X}^s} = \{ x_{(i)}^s\} _{i = 1}^{{N^s}}$ and the corresponding class labels ${{\cal Y}^s} = \{ y_{(i)}^s\} _{i = 1}^{{N^s}}$. The class labels span from 1 to $S$, ${y^s} \in {L^s} = \{ 1,...S\} $. 
In the FSZU task, the $\left| S \right|$ is same with GZSL. But the samples on every seen class are much less than GZSL, which contains ${{\cal X}^s} = \{ x_{(i)}^s\} _{i = 1}^{{N^f}}$ and ${{\cal Y}^s} = \{ y_{(i)}^s\} _{i = 1}^{{N^f}}$, where ${N^f} \ll {N^s}$.
For the test set which involves unseen classes, the GZSL is the same as the FSZU. Specifically, given another dataset  $\{ {{\cal X}^u},{{\cal Y}^u}\}$, on which the classes are related to the seen dataset.
% (e.g. all classes are from the same domain and have strong semantic relationships)
The dataset has $U$ unseen classes and consist of ${N^u}$ data instances ${{\cal X}^u} = \{ x_{(i)}^u\} _{i = 1}^{{N^u}}$ with corresponding labels ${{\cal Y}^u} = \{ y_{(i)}^u\} _{i = 1}^{{N^u}}$. The class labels thus range from $S+1$ to $S+U$, ${y^u} \in {L^u} = \{ S+1,...S+U\} $. The ${L^s} \cap {L^u} = \emptyset$. Each class is associated with a class-level semantic feature, which can be embedding and attribute. And the semantic information can be represented as ${\cal A} = \{ {a_{(k)}}\} _{k = 1}^{S + U}$. We denote ${\cal A}^s$ and ${\cal A}^u$ as the semantic features of seen and unseen classes. In this paper, the model training is achieved in two stages. We split the seen classes into source classes ($\{ {{\cal X}^{ss}},{{\cal Y}^{ss}}\}$, ${\cal A}^{ss}$) and target classes ($\{ {{\cal X}^{st}},{{\cal Y}^{st}}\}$, ${\cal A}^{st}$) in the first training stage. And we regard the seen classes as source classes and unseen classes as target classes in the second training stage.

\begin{figure*}
  \centering
  \includegraphics[width=170mm, height=80mm]{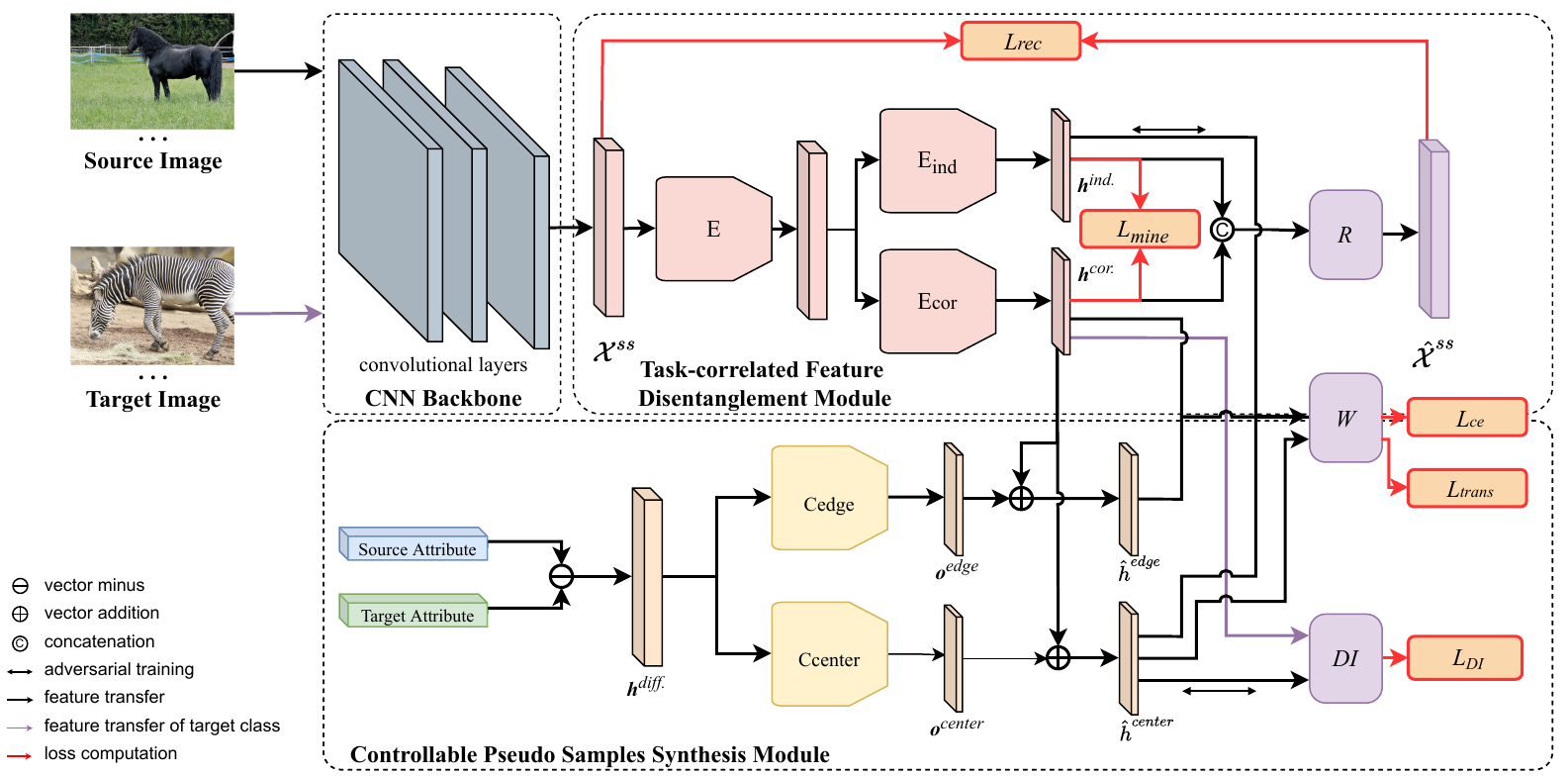}\\
  \caption{A schematic overview of TDCSS. The TDCSS is consist of two key component: (i) Feature Extractor Network $E$, Task-correlated  Network ${E_{cor}}$ and Task-independent Network ${E_{ind}}$, Reconstructor $R$, and $W$ are set for Task-correlated Feature Disentanglement Module. The representation disentanglement of task-correlated features ${h^{cor}}$ and task-independent features ${h^{ind}}$ is achieved by adversarial training. And the independence between ${h^{cor}}$ and ${h^{ind}}$ and the meaningfulness of two factors are guaranteed by mutual minimization ${L_{mine}}$ and reconstruction ${L_{rec}}$. (ii) The $E$, ${E_{cor}}$, $W$, Center Convert Net ${C_{center}}$, Edge Convert Nets ${C_{edge}}$, and Domain Identifier $DI$ are set for Controllable Pseudo Samples Synthesis Module. The inputs of convert nets are the target semantic features minus the source semantic features. And offsets ${o^{center}}$ and ${o^{edge}}$ add ${h^{cor}}$ of source images to synthesize center-pseudo samples ${{\hat h}^{center}}$ and edge-pseudo samples ${{\hat h}^{edge}}$ of target classes, respectively. The characteristics of ${{\hat h}^{center}}$ is guaranteed by the transfer loss ${L_{trans}}$ and the adversarial domain classification loss ${L_{DI}}$ with ${h^{cor}}$ of target visual features ${\cal X}^{st}$.}
  \label{fig:2}
  \vspace{-0.5cm}
\end{figure*}

\subsection{Overall Framework}
In this section, we present the details and the training strategy of TDCSS. The overall framework is illustrated in Figure \ref{fig:2}. There are two key components: (1) \textbf{Task-correlated feature disentanglement.} We first disentangle the confounding visual features ${{\cal X}^{ss}}$ of source classes to task-correlated features ${h^{cor}}$ and task-independent features ${h^{ind}}$ by adversarial training of domain adaptation and making sure that both of them are precise, meaningful and independent to each other. The disentangled task-correlated features ${h^{cor}}$ are then regarded as more reasonable representations for the sample synthesis. 
(2) \textbf{Controllable pseudo samples synthesis.} We use the task-correlated features ${h^{cor}}$ of source classes to add the Center Offset ${o^{center}}$ and the Edge Offsets ${o^{edge}}$ respectively, which can synthesize two types of pseudo samples that are center-pseudo samples ${{\hat h}^{center}}$ and edge-pseudo samples ${{\hat h}^{edge}}$. The offsets are outputted by convert networks. 
% Note that the inputs of convert nets are the difference between source semantic features ${\cal A}^{ss}$ and target semantic features ${\cal A}^{st}$. 

\subsection{Task-correlated Feature Disentanglement}
This module is consists of adversarial training, reconstruction, and mutual minimization.

\textbf{Adversarial Training.} We aim to disentangle the visual features into  ${h^{cor}}$ and ${h^{ind}}$ in an adversarial way. 

In the classification training step, we input visual features ${\cal X}^{ss}$ into Feature Extractor Network $E$, Task-correlated  Network ${E_{cor}}$ and Task-independent Network ${E_{ind}}$ to disentangle the vectors into two factors. Then we train $E$, ${E_{cor}}$ and ${E_{ind}}$ in the supervised ways by using the compatibility loss which associates the visual and the semantic. The compatibility score function is parameterized by $W$, and is typically formulated as the bilinear compatibility function: 
\begin
{equation}
\label{eq:single}
\tau _i^k = {h_i}W{a_k}
\end
{equation}
where the ${h_i}$ can be ${h^{cor}}$ or ${h^{ind}}$ of one sample after disentangling. And we further denote the attribute matrix:
\begin
{equation}
\label{eq:1}
\tau _i = {h_i}W{{\cal A}^s}
\end
{equation}
where the ${{\cal A}^s}$ should be ${\cal A}$ in the second training stage. We can consider ${\tau _i}$ as a classification score in the cross-entropy (CE) loss. So we further develop the compatibility loss function \cite{xie2019attentive}, which can be formulated as:
\begin
{equation}
\label{eq:2}
{L_{ce}} = \frac{1}{{{n_b}}}\sum\limits_{i = 1}^{{n_b}} {L({\tau _i},{y_i})} 
\end
{equation}
where the $L( \cdot )$ means the CE loss. The ${n_b}$ means the size of one batch.

In the adversarial training step, we fix the parameter $W$ of compatibility function and train $E$ and ${E_{ind}}$ to fool the classifier by minimizing the negative entropy of the predicted class distribution of ${h^{ind}}$ outputted by ${E_{ind}}$. 

\textbf{Reconstruction.} To guarantee the disentangled factors are meaningful, we reconstruct the confounding vector from them. Concretely, we concatenate ${h^{cor}}$ and ${h^{ind}}$, and then input it into Reconstructor $R$ to recover the confounding visual features ${\cal X}^{ss}$. Finally, the reconstruction loss function can be formulated as:
\begin
{equation}
\label{eq:rec}
{L_{rec}} = {\left\| {{{\hat {\cal X}^{ss}}} - {\cal X}^{ss}} \right\|_F^2}
\end
{equation}
where the ${\hat {\cal X}^{ss}}$ is the reconstructed vector of ${\cal X}^{ss}$, we use the ${L_{rec}}$ to train $R$

\textbf{Mutual Minimization.} We need to make sure two factors are independent of each other. Concretely, we minimize the mutual information \cite{belghazi2018mutual} between  ${h^{cor}}$ and ${h^{ind}}$ in an unsupervised way. The mutual information minimization loss function can be formulated as:
\allowdisplaybreaks
% \begin{eqnarray}
\label{eq:9}
% \begin{split}
\begin{align*}
    {L_{mine}} & = {\mathop{\rm Mine}\nolimits} (h^{cor},h^{ind})\\
    & = {\mathop{\rm H}\nolimits} ({h^{cor}}) - {\mathop{\rm H}\nolimits}({h^{cor}}|{h^{ind}})\\
    & = {\mathop{\rm H}\nolimits}({h^{ind}}) - {\mathop{\rm H}\nolimits}({h^{ind}}|{h^{cor}})\\
    & = \sum\limits_{{h^{cor}},{h^{ind}}} {p({h^{cor}},{h^{ind}})} \log \frac{{p({h^{cor}},{h^{ind}})}}{{p({h^{cor}})p({h^{ind}})}}
\end{align*}
% \end{split}
% \end{eqnarray}
% \begin
% {equation}
% \label{eq:3}
% &{L_{mine}} = {\mathop{\rm Mine}\nolimits} (h^{cor},h^{ind})
% \end
% {equation}
where ${\mathop{\rm H}\nolimits} ( \cdot )$ means the Shannon entropy and the ${\mathop{\rm H}\nolimits} ({h^{cor}}|{h^{ind}})$ means the conditional entropy of ${h^{cor}}$ given ${h^{ind}}$. The $p({h^{cor}},{h^{ind}})$ means the joint probability distribution of $({h^{cor}},{h^{ind}})$. We use the ${L_{mine}}$ to train $E$, ${E_{cor}}$ and ${E_{ind}}$.
% Minimize the mutual information means decreasing the correlation between the two factors.

\subsection{Controllable Pseudo Samples Synthesis}
This module is consists of pseudo sample synthesis and adversarial domain classification.

\textbf{Pseudo Sample Synthesis.} 
% We define ${{\hat h}^{edge}}$ and ${{\hat h}^{center}}$, which are distributed in edge and center of target classes respectively. 
Firstly, we use the Center Convert Net ${C_{center}}$ and the Edge Convert Nets ${C_{edge}}$ to generate ${o^{center}}$ and ${o^{edge}}$ respectively. The inputs of these networks are the difference between semantic features of target classes and that of source classes. Then the corresponding offsets are further added to ${h^{cor}}$ to synthesize ${{\hat h}^{center}}$ and ${{\hat h}^{edge}}$ of target classes respectively. This process can be formulated as:
\begin
{equation}
\label{eq:4}
{{\hat h}^{center}} = {h^{cor}} + {o^{center}}
\end
{equation}
\begin
{equation}
\label{eq:5}
{{\hat h}^{edge}} = {h^{cor}} + {o^{edge}}
\end
{equation}
where the $o^{center}$ and $o^{edge}$ can be further formulated as:
\begin
{equation}
\label{eq:7}
o_{ij}^{center} = {{\mathop{\rm C}\nolimits} _{center}}({a_i - a_j})
\end
{equation}
\begin
{equation}
\label{eq:8}
o_{ij}^{edge} = {{\mathop{\rm C}\nolimits} _{edge}}({a_i - a_j})
\end
{equation}
where $i$ and $j$ are the specific classes that from target classes and source classes respectively.

For different synthesizing samples, we have the following training process.
Firstly, we use ${{\hat h}^{center}}$ and ${{\hat h}^{edge}}$ to train the ${C_{center}}$ and the ${C_{edge}}$ with classification loss in Eq. \ref{eq:1} and Eq. \ref{eq:2} by labeling as target classes. Furthermore, to guarantee ${{\hat h}^{center}}$ distribute in the center of classes, we add the additional transfer loss to ${{\hat h}^{center}}$ to train the ${C_{center}}$. The transfer loss is based on Eq. \ref{eq:1} and Eq. \ref{eq:2} but using the soft labels that are computed by cosine similarity of semantic features between source classes and target classes. 
Then, we introduce the adversarial self-supervised \cite{kim2020adversarial} that based on ${{\hat h}^{edge}}$ to train the model. Concretely, for the ${{\hat h}^{edge}}$, the aforementioned process is similar to FSGM \cite{goodfellow2014explaining} algorithm and the offsets are similar to the perturbations in adversarial examples. 
We still label ${{\hat h}^{center}}$ as target classes to train the parameters $W$ of compatibility function by Eq. \ref{eq:1} and Eq. \ref{eq:2} while we label ${{\hat h}^{edge}}$ as source classes. In this way, we can strengthen the model's adversarial robustness and further contribute to the generalization of the model. 

% In this process, we also measure the pseudo samples' certainty by calculating their entropy like \cite{li2019leveraging} to screen fake samples.
% (the smaller the entropy, the higher the confidence). 
% For ${{\hat h}^{center}}$, we are more inclined to filter the pseudo samples with higher confidence and the screening is inverse for ${{\hat h}^{edge}}$.
% For ${{\hat h}^{edge}}$, we are more inclined to filter samples with lower confidence. For ${{\hat h}^{center}}$, we are more inclined to filter the pseudo samples with higher confidence.

\textbf{Adversarial Domain Classification.} We aim to synthesize pseudo samples more consistent with real samples which are achieved by adversarial domain classification. Specifically, we use the Domain Identifier $DI$, which takes ${{\hat h}^{center}}$ and ${h^{cor}}$ of ${\cal X}^{st}$ as input and output the domain label ${l_f}$ and $(1 - {l_f})$ respectively. The loss function can be formulated as:
\begin
{equation}
\label{eq:6}
{L_{DI}} =  - {l_f}\log ({l_f}) + (1 - {l_f})log(1 - {l_f})
\end
{equation}

Then the ${C_{center}}$ is trained by exchanging domain labels of real and pseudo samples to fool the $DI$. The ${{\hat h}^{edge}}$ distribute between source classes and target classes, so we have not taken them into domain classification.

% In a word, we use the ${{\hat h}^{edge}}$ to achieve adversarial self-supervised training and add the transfer loss and domain classification loss to ${{\hat h}^{center}}$ to make sure that the ${{\hat h}^{center}}$ distributes in the center of classes.

\subsection{Optimization and Unseen Samples Prediction}
Our model is trained with different losses iteratively. In the second training stage, we regard the seen classes as source classes and unseen classes as target classes. We use the transfer loss to finetune ${C_{center}}$ firstly and then synthesize ${{\hat h}^{center}}$ of target classes to finetune $W$ by Eq. \ref{eq:1} and Eq. \ref{eq:2}. The first training stage and the second training stage are running alternately in one epoch.

Once the model training is completed, we can project the visual features into semantic space and measure the similarity with the semantic features of all classes in the GZSL task. Specifically, to predict the class label, the location of the maximum compatibility score can be chosen as the predicted label:
\begin
{equation}
{y} = \arg \mathop {\max }\limits_{k \in {S+U}} \phi {({x})^T}a_k
\end
{equation}
where $\phi ( \cdot )$ includes the $E$, ${E_{cor}}$ and the $W$.

% \subsection{Formulation of FUZU}
% In the FSZU task, the $\left| S \right|$ is same with GZSL. But the samples on every seen class are much less than GZSL, which contains ${{\cal X}^s} = \{ x_{(i)}^s\} _{i = 1}^{{N^f}}$ and ${{\cal Y}^s} = \{ y_{(i)}^s\} _{i = 1}^{{N^f}}$, where ${N^f} \ll {N^s}$. For unseen classes, the class-level semantic representations, and the test sets, the GZSL and the FSZU are same. 

%-------------------------------------------------------------------------
\begin{table}
\small
\centering
\caption{The properties of datasets}
\label{tab:1}
\arrayrulecolor{black}
\setlength{\tabcolsep}{0.5pt}
{
\begin{tabular}{!{\color{black}\vrule}c!{\color{black}\vrule}c!{\color{black}\vrule}c!{\color{black}\vrule}c!{\color{black}\vrule}c!{\color{black}\vrule}}
\hline
\multirow{2}{*}{~}                 & \multicolumn{4}{c!{\color{black}\vrule}}{\textbf{ Dataset }}        \\ 
\cline{2-5}
& \textbf{ AWA1 } & \textbf{ AWA2 } & \textbf{ CUB } & \textbf{ FLO }  \\ 
\hline
\textbf{ \#Samples }                          
& 30475          
& 37322           
& 11788          
& 8189           
\\ 
\hline
\textbf{ \#Classes (train/test) }              
& 40/10          
& 40/10           
& 150/50         
& 82/20           
\\ 
\hline
\textbf{ Attributes }                       
& 85             
& 85              
& 1024          
& 1024            
\\ 
\hline
\makecell[c]{\textbf{Attribute value}\\\textbf{(Real or Boolean)}}
& both           
& both           
& Real           
& Real               
\\
\hline
\end{tabular}
}
\arrayrulecolor{black}
\end{table}

\begin{table}
\small
\centering
\caption{Difference between our and compared methods (The details include Non-Generative Model (NGM), Soft Label (SL), Overall Feature (OF), Representation Disentanglement (RD), Pseudo Sample Synthesis (PSS))}
\label{tab:diff}
\arrayrulecolor{black}
\setlength{\tabcolsep}{0.5pt}
{
\begin{tabular}{!{\color{black}\vrule}c!{\color{black}\vrule}c!{\color{black}\vrule}c!{\color{black}\vrule}c!{\color{black}\vrule}c!{\color{black}\vrule}c!{\color{black}\vrule}}
\hline
\textbf{ Models }
& \textbf{ NGM }
& \textbf{ SL } 
& \textbf{ OF } 
& \textbf{ RD } 
& \textbf{ PSS }  
\\ 
\hline
\textbf{DEM}
& \checkmark          
& $\times$           
& \checkmark           
& $\times$   
& $\times$                  
\\
\hline
\textbf{RELATION NET}
& \checkmark          
& $\times$          
& \checkmark           
& $\times$   
& $\times$                  
\\
\hline
\textbf{DCN}
& \checkmark          
& $\times$          
& \checkmark           
& $\times$  
& $\times$                  
\\
\hline
\textbf{TCN}
& \checkmark          
& \checkmark          
& \checkmark           
& $\times$   
& $\times$                  
\\
\hline
\textbf{ SP-AEN }                          
& \checkmark          
& $\times$          
& $\times$           
& \checkmark  
& $\times$   
\\ 
\hline
\textbf{ AREN+CS }              
& \checkmark          
& $\times$           
& $\times$            
& $\times$  
& $\times$             
\\ 
\hline
\textbf{AGZSL}
& \checkmark          
& $\times$         
& \checkmark           
& $\times$  
& \checkmark                 
\\
\hline
\textbf{ f-VAEGAN-D2 }                       
& $\times$           
& $\times$           
& \checkmark            
& $\times$   
& \checkmark            
\\ 
\hline
\textbf{DLFZRL}
& $\times$           
& $\times$          
& \checkmark           
& \checkmark  
& $\times$                 
\\
\hline
\textbf{TDCSS}
& \checkmark          
& \checkmark          
& \checkmark           
& \checkmark  
& \checkmark                 
\\
\hline
\end{tabular}
}
\arrayrulecolor{black}
\vspace{-0.3cm}
\end{table}

\begin{table*}
\small
\centering
\caption{Performance (in \%) comparisons for GZSL in terms of unseen accuracy (u), seen accuracy (s), and their harmonic mean (H).} 
% $*$ indicates the results of this model obtained by ourselves with the codes released by the authors.
\label{tab:2}
\arrayrulecolor{black}
\begin{tabular}{!{\color{black}\vrule}c!{\color{black}\vrule}c!{\color{black}\vrule}c!{\color{black}\vrule}c!{\color{black}\vrule}c!{\color{black}\vrule}c!{\color{black}\vrule}c!{\color{black}\vrule}c!{\color{black}\vrule}c!{\color{black}\vrule}c!{\color{black}\vrule}c!{\color{black}\vrule}c!{\color{black}\vrule}c!{\color{black}\vrule}} 
\hline
\multirow{2}{*}{\textbf{Methods}} &
\multicolumn{3}{c!{\color{black}\vrule}}{\textbf{AWA1}} & \multicolumn{3}{c!{\color{black}\vrule}}{\textbf{AWA2}} & \multicolumn{3}{c!{\color{black}\vrule}}{\textbf{CUB}} & \multicolumn{3}{c!{\color{black}\vrule}}{\textbf{FLO}}  \\ 
\cline{2-13}
& u        & s        & \textbf{H}                        
& u        & s        & \textbf{H}                        
& u         & s        & \textbf{H}                      
& u        & s        & \textbf{H}                        \\ 
\hline
\textbf{DEM}  \cite{zhang2017learning}          
& 32.8      & 84.7    & 47.3
& 30.5      & 86.4    & 45.1
& 19.6      & 57.9    & 29.2
& 57.2      & 67.7    & 62.0   \\ 
\hline
\textbf{RELATION NET} \cite{sung2018learning} 
& 31.4      & 91.3    & 46.7
& 30.0      & 93.4    & 45.3
& 38.1      & 61.1    & 47.0
& 50.8      & 88.5    & 64.5   \\ 
\hline
\textbf{DCN} \cite{liu2018generalized}          
& -        & -        & -                        
& 25.5     & 84.2     & 39.1
& 28.4     & 60.7      & 38.7                   
& -        & -        & -                        \\ 
\hline
\textbf{TCN} \cite{jiang2019transferable}          
& -        & -        & -                        
& 61.2     & 65.8     & 63.4
& 52.6     & 52.0     & 52.3
& -        & -        & -                        \\
\hline
\textbf{SP-AEN} \cite{chen2018zero}    
& -     & -     & -                     
& 23.3     & 90.0     & 37.1                     
& 34.7      & 70.6     & 46.6                   
& -        & -        & -                        \\ 
\hline
\textbf{AREN+CS} \cite{xie2019attentive}    
& -     & -     & -                     
& 54.7   & 79.1  & 64.7
& 63.2   & 69.0  & 66.0
& -        & -        & -                        \\ 
\hline
\textbf{AGZSL} \cite{chou2020adaptive}         
& -        & -        & -                        
& 46.6     & 74.2     & 57.3
& 42.1    & 48.1     & 44.9
& -        & -        & -                        \\ 
\hline
\textbf{f-VAEGAN-D2} \cite{xian2019f}  
& 57.6     & 70.6     & 63.5                     
& -        & -        & -                        
& 48.4      & 60.1     & 53.6                   
& 56.8     & 74.9     & 64.6                     \\ 
\hline
\textbf{DLFZRL} \cite{tong2019hierarchical}     
&  -     & -     & 61.2                     
&  -     & -     & 60.9                     
& -      & -     & 51.9                   
& -        & -        & -                        \\ 
\hline
\textbf{TDCSS}    
& 54.4        & 69.8        & 60.9                       
& 59.2        & 74.9     & \textbf{66.1}                     
& 44.2         & 62.8        & 51.9                     
& 54.1        & 85.1        & \textbf{66.2}                        \\
\hline
\end{tabular}
\arrayrulecolor{black}
\vspace{-0.3cm}
\end{table*}

\section{Experiments}
\subsection{Experiments Setting}
\textbf{Datasets.} 
% Among the widely used datasets for GZSL image classification, 
We selected four popular datasets which are Animal with Attribute (AWA1) \cite{lampert2013attribute}, Animal with Attribute2 (AWA2) \cite{xian2018zero}, Caltech-UCSD Birds-200-2011 (CUB) \cite{wah2011caltech} and Oxford 102 flowers (FLO) \cite{nilsback2008automated}. AwA1 and AwA2 are coarse-grained while others are fine-grained. The semantic features of CUB and FLO are from the CNN-RNN features \cite{yu2020episode,reed2016learning}. Our dataset split is under the PS setting \cite{xian2018zero}. The details are presented in Table \ref{tab:1}.
% The details of the datasets are presented in Table \ref{tab:1}.
% \usepackage{colortbl}

\textbf{Evaluation Metric.} Average Class Accuracy (ACA) is adopted as the evaluation metric in the GZSL and FSZU tasks. We use the average per-class top-1 accuracy of unseen classes $U$ and seen classes $S$ to calculate the harmonic mean $H$:
\begin
{equation}
H = \frac{{(2 \times U \times S)}}{{U + S}}
\end
{equation}

% \subsection{Experiments Setting}
\textbf{Comparison Methods.} Since our model is a non-generative model, we mainly compare our proposed methods against current non-generative models.
% on all four aforementioned ZSL datasets. 
The main differences between compared methods and our TDCSS are shown in Table \ref{tab:diff}. 

\begin{table}
\small
\centering
\caption{Ablation study (in \%) of the TDCSS components on AWA2 datasets in terms of unseen accuracy (u), seen accuracy (s), and their harmonic mean (H).}
\label{tab:3}
\arrayrulecolor{black}
\begin{tabular}{!{\color{black}\vrule}c!{\color{black}\vrule}c!{\color{black}\vrule}c!{\color{black}\vrule}c!{\color{black}\vrule}} 
\hline
\multirow{2}{*}{\textbf{Setting}} &
\multicolumn{3}{c!{\color{black}\vrule}}{\textbf{AWA2}}  \\ 
\cline{2-4}
& u      
& s      
& \textbf{H}                             \\ 
\hline
\textbf{TDCSS w/o TFD}
& 52.7      
& 74.4      
& 61.7                             
\\ 
\hline
\textbf{TDCSS w/o EPS}     
& 44.5      
& 71.9      
& 55.0                             
\\ 
\hline
\textbf{TDCSS w/o CPS}  
& 34.9      
& 79.3      
& 48.4                             
\\ 
\hline
\textbf{TDCSS}                     
& 59.2      
& 74.9      
& \textbf{66.1}                       
\\
\hline
\end{tabular}
\arrayrulecolor{black}
\vspace{-0.3cm}
\end{table}

\begin{figure*}
  \centering
  \includegraphics[width=180mm, height=48mm]{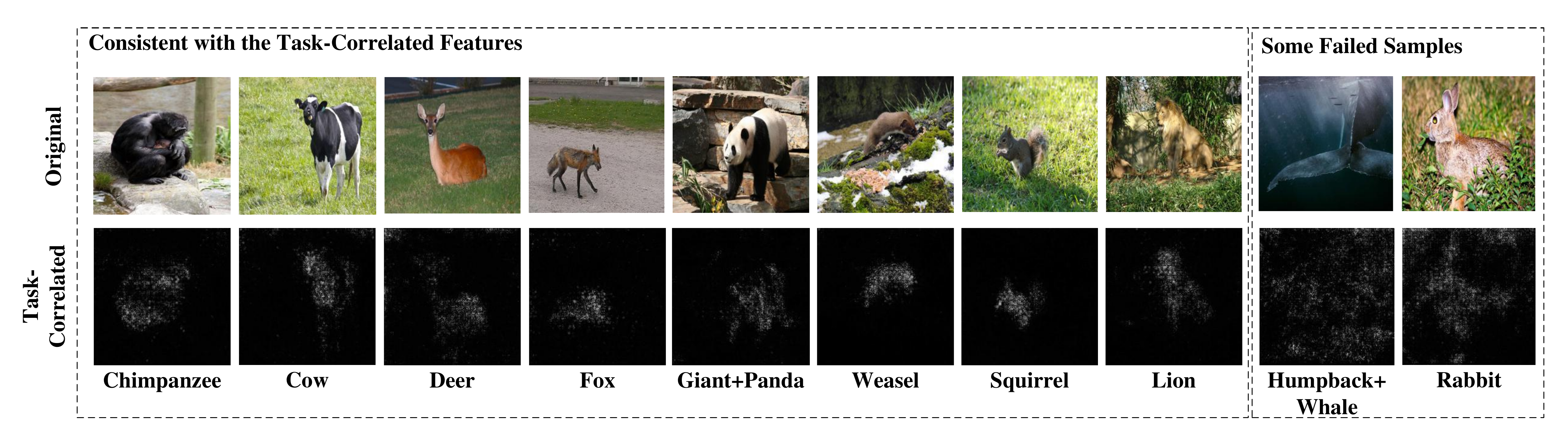}\\
  \caption{Visualization of task-correlated features that from AWA2 dataset by saliency maps}
  \label{fig:semantic_visulization}
  \vspace{-0.3cm}
\end{figure*}

\begin{figure}
  \centering
  \includegraphics[width=65mm, height=65mm]{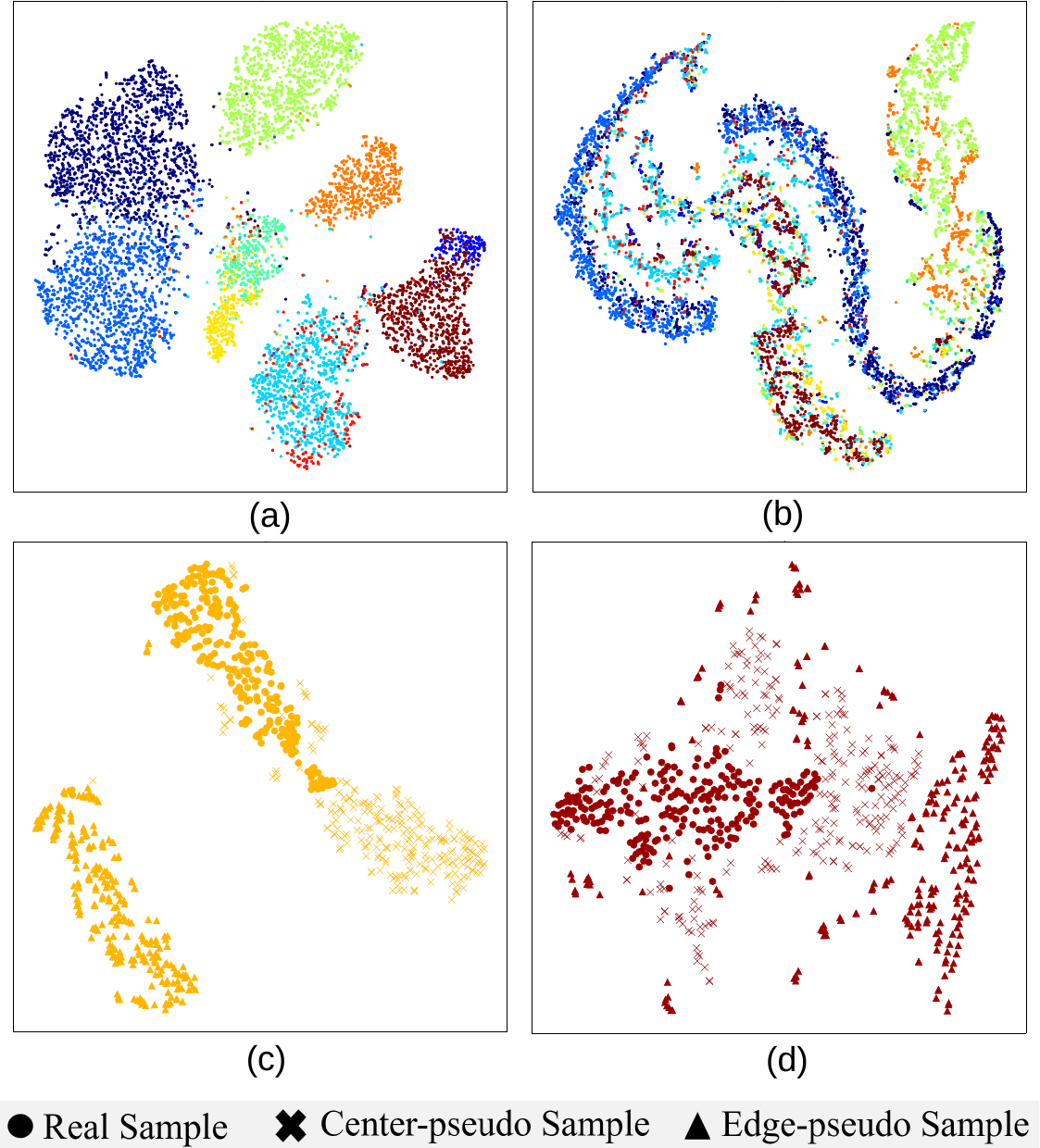}\\
  \caption{The t-SNE visual results of samples distributions on AWA2 dataset. (a) The task-correlated features ${h^{cor}}$. (b) The task-independent features ${h^{ind}}$. (c) \& (d) The distributions of the real samples and different types of pseudo samples.}
  \label{fig:4}
  \vspace{-0.5cm}
\end{figure}

\textbf{Implementation Details.} We utilize the 2048D visual features extracted by pre-trained ResNet-101 \cite{he2016deep}. The $E$, ${E_{cor}}$ (${E_{ind}}$), and $DI$ consist of two-layer fully connected (FC) neural networks, in which the output units are 1800, 1024, and 2 respectively. The $C$, $W$, and $R$ are three-layer FC neural networks, in which the hidden units are 1024, 512, and 1800. The output units of $C$ are 1024. We use the LeakyReLU as the activation function for $DI$ while the ReLU for others. Our model is implemented with PyTorch and optimized by ADAM optimizer. We set the learning rate as 2e-4 in the first training stage and 1/10 in the second training stage, epoch as 1500 for most. And in every epoch, we iterate 30 batches in the first training stage and 10 in the second training stage. Because of the limitation of sample size on every class, the batch size is 32 for the FLO dataset and 64 for the other datasets. For source/target split in the first stage, we set the number of target classes to 2 for CUB and 1 for others.

\subsection{Evaluations in GZSL Setting}
The classification performances in the GZSL tasks are shown in Table \ref{tab:2}. We observe that the TDCSS achieves competitive results on four datasets. 

Compared with non-generative models, the $H$ value of our model increases from 47.3\% to 60.9\% on AWA1, from 64.7\% to 66.1\% on AWA2, and from 64.5\% to 66.2\% on FLO. 
% And on the CUB dataset, the model has comparable results compared with the existed models. 
Specifically, the TCN contains the soft labels to quantify the transfer process in GZSL, which we add to ${{\hat h}^{center}}$ synthesis process. It can be concluded that our model achieves improvement in addition to the transfer loss. 
The SP-AEN and AREN are based on tensor-level features. For the SP-AEN that tries to disentangle features, the experimental results prove our model is more effective. For the AREN+CS, our model is still competitive except on CUB dataset. However, the mechanism of calibrated stacking (CS) \cite{chao2016empirical} helps the AREN achieve a great improvement in the GZSL task. But it is a post-processing operation and very susceptible to the influence of the parameter values that are manually set by the researchers. Our model does not use the CS but with comparable results, which shows TDCSS is more robust.  
For the AGZSL synthesized samples but with no representation disentanglement, which proves the effectiveness of representation disentanglement of our model. 

Compared with generative models, the $H$ value of our model increases from 60.9\% to 66.1\% on AWA2 and from 64.6\% to 66.2\% on FLO. However, the pseudo samples synthesized by our model have certain characteristics, which can further explore the role of different types of pseudo samples in GZSL knowledge transfer. The further detailed experiments are shown in section \ref{sec: ablation}. The DLFZRL disentangles features by generative networks, our model also achieves comparable results generally but in a simpler concept and method.

\subsection{Ablation Study}
\label{sec: ablation}
% In this part, we take the AWA2 dataset into the following ablation analysis. 
We take the AWA2 dataset into the ablation analysis and aim to demonstrate that the main components of the TDCSS both contribute to the final performance. We also observe the role of different types of pseudo samples in the knowledge transfer of the GZSL task. The best performance is achieved when TFD (Task-correlated Feature Disentanglement), EPS (Edge-Pseudo Samples) and CPS (Center-Pseudo Samples) are both applied. We have the following main findings:

% Besides the basic adversarial training of domain adaptation, the TDCSS mainly consists of task-correlated feature disentanglement and two types of pseudo samples synthesis, so we further observe the following three cases: (1) model w/o task-correlated feature disentanglement. (2) model w/o edge-pseudo samples. (3) model w/o center-pseudo samples. The results are presented in Table \ref{tab:3}. We can observe the model with all components mostly achieves the best performances. Concretely, we have the following main findings:

(1) Compared with the model w/o TFD, it can be shown that disentanglement has little effect on the recognition of seen classes in the model, but has a greater impact on unseen classes. It proves that the disentanglement module, which extracts more consistent visual features with class semantics, is of great help to the knowledge transfer from the seen class to the unseen class. 

(2) Compared with the model w/o EPS, the experiments show that our whole model has a certain improvement in the accuracy of the seen and unseen classes, indicating that ${{\hat h}^{edge}}$ with adversarial self-supervised training contribute to the consistency between ${h^{cor}}$ and class semantics, and further improving the robustness and generalization of the model.

(3) Compared with the model w/o CPS, on the one hand, the precision on seen classes of the whole TDCSS is worse than it. It has demonstrated that ${{\hat h}^{edge}}$ can further perfect the classification boundary for seen classes. On the other hand, the precision on unseen classes has significantly declined, it has proved ${{\hat h}^{center}}$ play a key role in the knowledge transfer from the seen classes to the unseen classes in the GZSL task.

\begin{table*}
\small
\centering
\caption{Performance (in \%) comparisons for FSZU in terms of unseen accuracy (u), seen accuracy (s), and their harmonic mean (H).}
\label{tab:5}
\arrayrulecolor{black}
\begin{tabular}
{!{\color{black}\vrule}c!{\color{black}\vrule}c!{\color{black}\vrule}c!{\color{black}\vrule}c!{\color{black}\vrule}c!{\color{black}\vrule}c!{\color{black}\vrule}c!{\color{black}\vrule}c!{\color{black}\vrule}c!{\color{black}\vrule}c!{\color{black}\vrule}c!{\color{black}\vrule}c!{\color{black}\vrule}c!{\color{black}\vrule}} 
\hline
\multirow{2}{*}{\textbf{AWA2}} &
\multicolumn{3}{c!{\color{black}\vrule}}{\textbf{All data}} & \multicolumn{3}{c!{\color{black}\vrule}}{\textbf{Num = 10}} & \multicolumn{3}{c!{\color{black}\vrule}}{\textbf{Num = 5}} & \multicolumn{3}{c!{\color{black}\vrule}}{\textbf{Num = 2}} \\ 
\cline{2-13}
 & u        & s        & \textbf{H}                        
 & u        & s        & \textbf{H}                       
 & u         & s        & \textbf{H}
 & u         & s        & \textbf{H}\\ 
\hline
\textbf{Disentangled-VAE} \cite{li2021generalized}   
& 50.9     & 79.8     & 62.2                     
& 50.8     & 64.7     & 56.9                    
& 39.2      & 58.1     & 46.8
& 29.8      & 39.7     & 34.1
\\ 
\hline
\textbf{SDGZSL} \cite{chen2021semantics}   
& 74.4	& 63.6	& 68.6                     
& 47.1	& 53.3	& 50.0                   
& 25.7	& 56.1	& 35.3
& 6.8	   & 39.9	    & 11.6
\\
\hline
\textbf{AGZSL} \cite{chou2020adaptive}  
& 46.6	& 74.2	& 57.3	
& 18.3	& 81.1	& 29.9	
& 14.7	& 71.9	& 24.5	
& 13.2	& 60.1	& 21.6
\\ 
\hline
\textbf{TDCSS}   
& 59.2	& 74.9	& 66.1                     
& 56.3	& 60.9	& \textbf{58.5}                     
& 49.0	& 69.1	& \textbf{57.3}  
& 39.8	& 61.8	& \textbf{48.4}
\\ 

\hline\hline
\multirow{2}{*}{\textbf{CUB}} &
\multicolumn{3}{c!{\color{black}\vrule}}{\textbf{All data}} & \multicolumn{3}{c!{\color{black}\vrule}}{\textbf{Num = 10}} & \multicolumn{3}{c!{\color{black}\vrule}}{\textbf{Num = 5}} & \multicolumn{3}{c!{\color{black}\vrule}}{\textbf{Num = 2}} \\ 
\cline{2-13}
 & u        & s        & \textbf{H}                        
 & u        & s        & \textbf{H}                       
 & u         & s        & \textbf{H}
 & u         & s        & \textbf{H}\\ 
 \hline
\textbf{Disentangled-VAE} \cite{li2021generalized}   
& 52.1     & 54.2     & 53.1                    
& 44.7      & 47.5     & 46.1
& 38.9      & 39.8     & 39.4
& 37.9     & 26.6     & 31.2    
\\ 
\hline
\textbf{SDGZSL} \cite{chen2021semantics}   
& 61.2	& 65.3	& 63.2                     
& 39.4	& 59.1	& 47.3                    
& 26.2	& 55.6	& 35.6
& 7.5	& 47.0	& 12.9
\\ 
\hline
\textbf{AGZSL} \cite{chou2020adaptive}  
& 42.1	& 48.1	& 44.9	
& 29.3	& 42.9	& 34.8	
& 21.7	& 36.1	& 27.1	
& 13.7	& 26.8	& 18.1
\\ 
\hline
\textbf{TDCSS}   
&  44.2     & 62.8   & 51.9                    
& 40.1     & 54.5     & 46.2                  
& 43.4      & 45.5     & \textbf{44.4}   
& 34.5	   & 38.1	  & \textbf{36.2}
\\ 

\hline
\end{tabular}
\arrayrulecolor{black}
\vspace{-0.3cm}
\end{table*}

\subsection{Qualitative Analysis }
We take the AWA2 dataset into the following qualitative analysis. 

\textbf{Task-correlated features visualization}
We visualize ${h^{cor}}$ and ${h^{ind}}$ in Figure \ref{fig:4} (a) and (b) to validate the properties of the disentanglement. It shows that ${h^{cor}}$ are much more discriminative than ${h^{ind}}$. But ${h^{ind}}$ remain some discriminative, which we guess that some characteristics are not annotated in semantics. We further visualize ${h^{cor}}$ by saliency maps \cite{simonyan2014deep} which compute the gradient of the output of ${E_{cor}}$ against the original image inputted in the backbone network. The results are shown in Figure \ref{fig:semantic_visulization}. We can observe that the saliency maps focus on the task-correlated information, especially the objects. And the task-independent information is effectively filtered. But there are also some failed samples that animals blend with their surroundings. 
% In general, these results show that the task-correlated feature disentanglement module of TDCSS is effective.

\textbf{Distribution visualization of pseudo samples}
To demonstrate that our method can synthesize two types of pseudo samples effectively, we randomly select two classes and visualize the distributions of part samples. As Figure \ref{fig:4} (c) and (d) shows, we can observe that the distributions of ${{\hat h}^{center}}$ are closer to real samples than that of ${{\hat h}^{edge}}$ on the whole, which are consistent with their characteristics.
% As Fig. \ref{fig:3} shows, we can observe that the synthesized pseudo samples are consistent with their characteristics. Specifically, (1) the center-pseudo samples are closer to the real samples than edge-pseudo samples. 
% It can be seen in Fig. \ref{fig:4} more clearly, which we randomly select samples from individual classes respectively.  (2) The size of edge-pseudo samples is the same as that of center-pseudo samples. The latter one is distributed in one cluster but the first one is not. This phenomenon can be explained as the edge-pseudo samples are distributed among different classes, which do not follow strict cluster distribution. So the edge-pseudo samples of different classes present mixed distribution, which are consistent with our expectations. Both of these phenomenons can demonstrate the effectiveness of the controllable pseudo samples synthesis module.

% ----------------------------------------------------------------
\begin{figure}
  \centering
  \includegraphics[width=80mm, height=32mm]{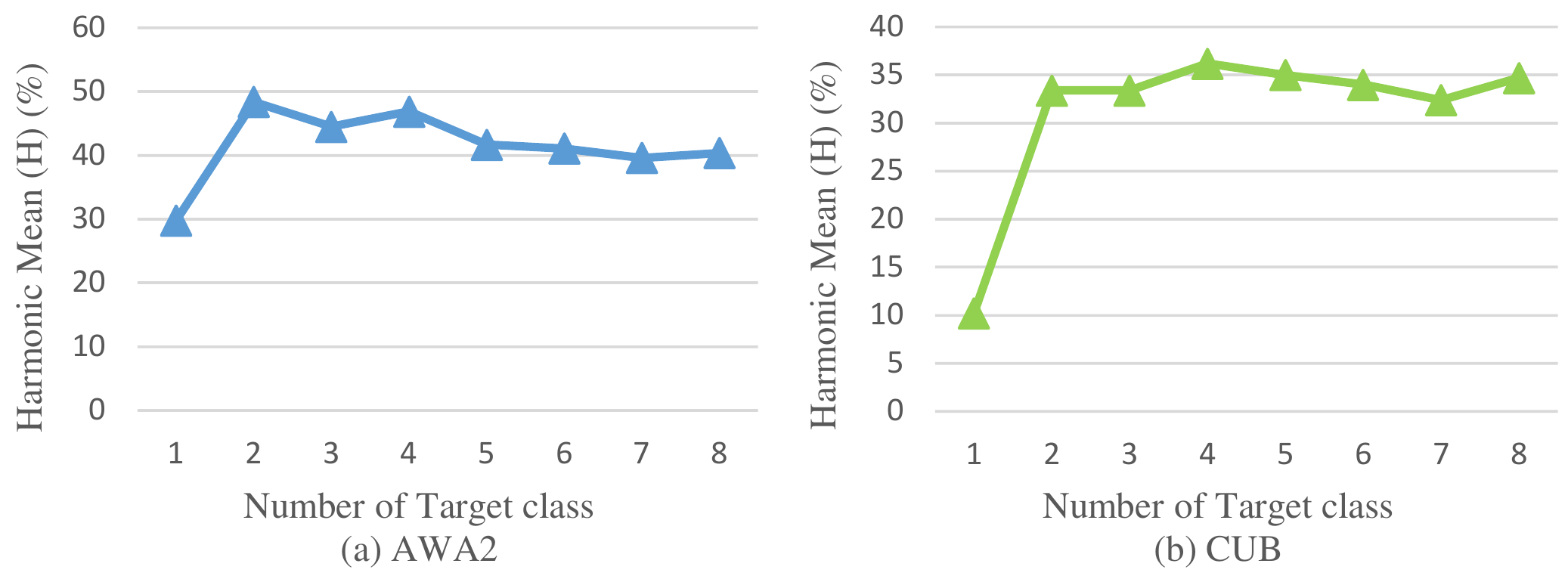}\\
  \caption{The experimental results under different numbers of target classes when the sample size of each class is 2.}
  \label{fig:5}
  \vspace{-0.5cm}
\end{figure}

\subsection{Evaluations in FSZU Setting}
% In this section, we will discuss the performance of models in the FSZU task. 
For compared methods, we select two generative models that disentangle the overall representations into two factors by random permuting. The Disentangled-VAE \cite{li2021generalized} consists of two parallel VAEs and each with two branches. 
% It trains the final classifier with the hidden layer representations. 
The SDGZSL \cite{chen2021semantics} consists of VAE, AE, and the RELATION NET \cite{sung2018learning}, while the VAE is used for data enhancement. We also select the non-generative AGZSL that synthesizes pseudo samples in mixup interpolation for comparison.

For experimental settings, We select AWA2 and CUB datasets, that appeared simultaneously in the Disentangled-VAE, SDGZSL, and AGZSL. And we reduce the sample size of seen classes (the sizes are set to 10, 5, and 2) to stimulate the FSZU. It should be pointed out that the Disentangled-VAE is reproduced by us based on Python 3.6 and Pytorch 1.0.1. For SDGZSL and AGZSL, we use the codes that have been released on Github. The performances in the FSZU tasks are shown in Table \ref{tab:5}. 

% 开始实验分析
Compared with generative models, we achieve 22.4\% improvement in $H$ value on average for AWA2 and 8.8\% for CUB. These experimental results reflect that our model can still work effectively in the new task. The generative model, especially SDGZSL, has excellent performances in GZSL but degrade sharply with the decrease in sample size, which shows more non-robust compared with our model. And our model can synthesize more diverse pseudo samples based on the limited seen class samples in a non-generative way. 

Compared with the non-generative model that synthesizes pseudo samples, we achieve 117.9\% improvement in $H$ value on average for AWA2 and 65.5\% for CUB. It shows that the representation disentanglement before synthesizing samples is reasonable and important. The results in the FSZU also show that the mixup interpolation of the AGZSL will lead the diversity of the synthesis samples to decrease sharply as the sample size decreases.

In the task of FSZU, we can improve the model's performance by increasing the sample size of the target class with increasing the number of target classes. The results are shown in Figure \ref{fig:5}. It shows that within a certain range, the accuracy will increase as the number of target classes increases, but too many target classes will cause the performance to decrease. It is a corollary that too many synthesized pseudo samples would be the leading data in the training process and further distraction the model from recognizing real samples.

\section{Conclusions}
In this paper, we propose a non-generative model, TDCSS, to perform the task-correlated feature disentanglement and diversity pseudo samples synthesis in the GZSL and the FSZU tasks. For disentanglement, the TDCSS uses the adversarial training of domain adaptation to achieve it. For synthesis, the TDCSS synthesizes diverse pseudo samples with certain characteristics. The above mechanisms make our model achieve competitive performances in different tasks, and help people intuitively understand the role of different types of pseudo-samples in the knowledge transfer of ZSL.

% In this paper, we first propose a non-generative model to perform the task-correlated feature disentanglement and diversity pseudo samples synthesis in the challenging GZSL task. Furthermore, we propose a more real new task named FSZU. Firstly, we disentangle the information of original visual features and make the task-correlated features more consistent with the class semantic features. Secondly, we synthesize two types of pseudo samples to predict the class distributions. The edge-pseudo samples can be seen as the adversarial examples in the feature-level and further contribute to the generalization and robustness of our model. The center-pseudo samples are the key in the knowledge transfer. Finally, the experiments in the GZSL task and the FSZU task have proved our model achieves competitive results on standard benchmarks.

\section*{Acknowledgments}
This work is supported by the National Key R\&D Program of China (Grant No. 2018AAA0100604), the National Natural Science Foundation of China (Grant No. 61832004, 61632002), and Beijing Natural Science Foundation (No.JQ20023).

%%%%%%%%% REFERENCES
{\small
\bibliographystyle{ieee_fullname}
\bibliography{egbib2}
}

\end{document}